\documentclass{article}

\usepackage{arxiv}

\usepackage[utf8]{inputenc}
\usepackage[T1]{fontenc}
\usepackage{hyperref}
\usepackage{url}
\usepackage{booktabs}
\usepackage{amsfonts}
\usepackage{amsmath}
\usepackage{graphicx}
\usepackage{xcolor}
\usepackage{subcaption}
\usepackage{multirow}
\usepackage{array}

\title{PrivHAR-Bench: A Graduated Privacy Benchmark Dataset for Video-Based Action Recognition}

\author{
	Samar Ansari\\
	School of Computer and Engineering Sciences\\
	University of Chester\\
	Chester, CH1 4BJ, United Kingdom \\
	\texttt{m.ansari@chester.ac.uk}
}

\begin{document}

\maketitle

%!TEX root = Privacy_Dataset_Paper.tex

\begin{abstract}

Existing research on privacy-preserving Human Activity Recognition (HAR) typically evaluates methods against a binary paradigm: clear video versus a single privacy transformation. This limits cross-method comparability and obscures the nuanced relationship between privacy strength and recognition utility. We introduce \textit{PrivHAR-Bench}, a multi-tier benchmark dataset designed to standardize the evaluation of the \textit{Privacy-Utility Trade-off} in video-based action recognition. PrivHAR-Bench applies a graduated spectrum of visual privacy transformations: from lightweight spatial obfuscation to cryptographic block permutation, to a curated subset of 15 activity classes selected for human articulation diversity. Each of the 1,932 source videos is distributed across 9 parallel tiers of increasing privacy strength, with additional background-removed variants to isolate the contribution of human motion features from contextual scene bias. We provide lossless frame sequences, per-frame bounding boxes, estimated pose keypoints with joint-level confidence scores, standardized group-based train/test splits, and an evaluation toolkit computing recognition accuracy and privacy metrics. Empirical validation using R3D-18 demonstrates a measurable and interpretable degradation curve across tiers, with within-tier accuracy declining from 88.8\% (clear) to 53.5\% (encrypted, background-removed) and cross-domain accuracy collapsing to 4.8\%, establishing PrivHAR-Bench as a controlled benchmark for comparing privacy-preserving HAR methods under standardized conditions. The dataset, generation pipeline, and evaluation code are publicly available.

\end{abstract}

%!TEX root = Privacy_Dataset_Paper.tex

\section{Introduction}
\label{sec:introduction}

Video-based Human Activity Recognition (HAR) has achieved remarkable accuracy on standard benchmarks \cite{soomro2012ucf101, kay2017kinetics, shahroudy2016ntu}, driven by advances in 3D convolutional networks \cite{tran2015learning, carreira2017quo}, vision transformers \cite{bertasius2021space}, and large-scale pre-training \cite{tong2022videomae}. However, deploying these systems in privacy-sensitive environments, such as elderly care facilities, retail spaces, and public infrastructure, requires reconciling two competing objectives: preserving sufficient visual information for accurate activity classification while destroying enough information to prevent the identification of individuals.

A growing body of work addresses this tension through visual privacy transformations applied to video data prior to recognition. Proposed methods range from conventional obfuscation techniques such as blurring and pixelation \cite{tu2021toward, al2020modeling}, to structural representations like skeleton extraction \cite{gao2024privacy, yan2018spatial}, to cryptographic approaches including pixel-level encryption and block scrambling \cite{ishikawa2024learnable, qi2023privacy, ramasamy2019image}. Each method occupies a different point on the spectrum between privacy protection and recognition utility.

Despite this proliferation of methods, the field lacks a standardized benchmark for comparing them. Current evaluation practices suffer from three structural problems:

\paragraph{Binary evaluation.} Most studies compare their method against a single baseline: unmodified video versus their proposed transformation \cite{wu2020privacy, li2023stprivacy, kumawat2022privacy}. This binary framing reveals whether a method ``works'' but provides no insight into \textit{how much} privacy can be traded for \textit{how much} utility. Without intermediate reference points, it is impossible to determine whether a 5\% accuracy drop reflects strong privacy or a weak recognition model.

\paragraph{Context bias contamination.} A well-documented limitation of HAR datasets is that models frequently learn to classify actions from background context rather than human motion \cite{rastegar2024background, zhou2025seeing}. A boxing ring predicts boxing; a kitchen predicts cooking. When privacy transformations are applied only to the human region of interest (ROI), background context remains fully visible, potentially allowing models to bypass the privacy transformation entirely. No existing privacy-preserving HAR evaluation systematically controls for this confound.

\paragraph{Non-comparable evaluation protocols.} In the absence of standardized data splits, resolution, temporal windows, and evaluation metrics, results reported across different studies cannot be directly compared \cite{abbasi2024trading}. Differences in random splits, input preprocessing, and metric computation introduce uncontrolled variance that obscures genuine methodological differences.

%% --- FIGURE: Motivation figure showing the same action under different privacy tiers ---
%\begin{figure}[t]
%    \centering
%    \includegraphics[width=0.99\textwidth]{tier_overview.png}
%    \caption{A single frame from the PrivHAR-Bench dataset shown across all privacy tiers. From left to right: Original (no privacy), Tier~1 Blur ($\sigma=15$), Tier~2 Edge (Canny), Tier~3 AES scramble at block sizes $B=4$, $B=8$, and $B=16$, and the background-removed (NoBG) variant of $B=8$. As tier level increases, spatial identity features are progressively destroyed while gross motion structure is preserved at varying degrees.}
%    \label{fig:tier_overview}
%\end{figure}

\begin{figure*}[t]
	\centering
	\includegraphics[width=0.98\textwidth]{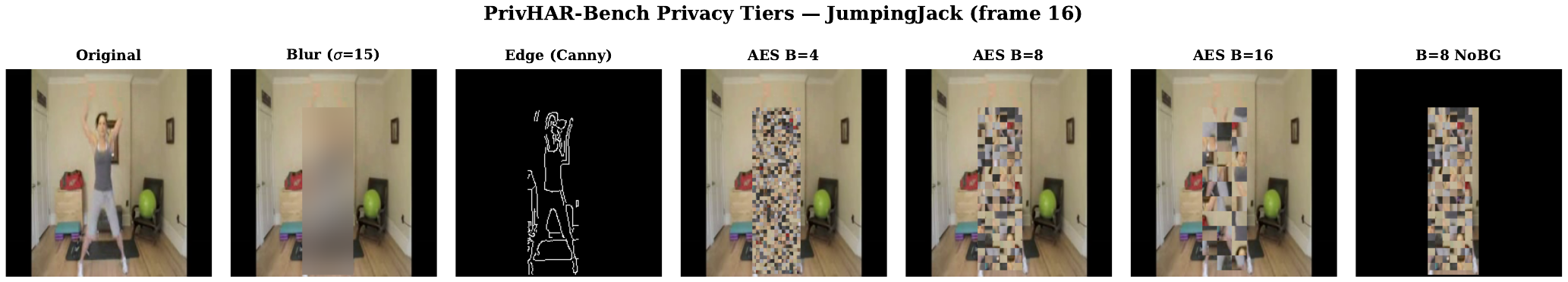}
	\caption{A single frame from the PrivHAR-Bench dataset shown across selected privacy tiers. From left to right: Original (no privacy), Tier 1 Blur ($\sigma=15$), Tier 2 Edge (Canny), Tier 3 AES scramble at block sizes $B=4$, $B=8$, and $B=16$, and the background-removed (NoBG) variant of $B=8$. As tier level increases, spatial identity features are progressively destroyed while gross motion structure is preserved at varying degrees.}
	\label{fig:tier_overview}
\end{figure*}

We introduce \textit{PrivHAR-Bench}, a multi-tier dataset that addresses these three problems simultaneously. PrivHAR-Bench provides:

\begin{enumerate}
    \item \textit{A graduated privacy spectrum.} Every video in the dataset is distributed across multiple parallel privacy tiers, ranging from Gaussian blur (low privacy, high utility) through structural edge extraction (medium privacy) to cryptographic block scrambling at multiple granularities (high privacy). This enables researchers to plot their model's accuracy as a continuous function of privacy strength, rather than reporting a single binary comparison.

    \item \textit{Systematic context bias control.} For every encrypted video, a background-removed (NoBG) variant is provided in which all non-ROI pixels are set to zero. This forces models to demonstrate that recognition performance derives from the transformed human region, not from environmental cues.

    \item \textit{A fixed evaluation protocol.} PrivHAR-Bench includes standardized group-based train/test splits, a prescribed temporal window of 32 frames, lossless frame storage to prevent codec-induced degradation of privacy transformations, and an evaluation toolkit that computes Top-1 accuracy, SSIM, PSNR, and a composite Privacy-Utility score.
\end{enumerate}

The dataset is constructed from a curated subset of 15 classes drawn from UCF101 \cite{soomro2012ucf101}, selected for diversity of human body articulation and minimal dependence on background context or non-human objects. All data is distributed as lossless PNG frame sequences to preserve the mathematical properties of the applied transformations. The complete generation pipeline, pinned dependencies, and deterministic seeds are released alongside the dataset to enable exact reproduction.

The remainder of this paper is organized as follows. Section~\ref{sec:related_work} reviews related work in privacy-preserving HAR and existing benchmarks. Section~\ref{sec:benchmark_design} describes the design principles and tier structure. Section~\ref{sec:dataset_generation} details the generation pipeline. Section~\ref{sec:evaluation_protocol} defines the evaluation protocol and metrics. Section~\ref{sec:experiments} presents baseline experiments. Section~\ref{sec:limitations} discusses limitations and future work, and Section~\ref{sec:conclusion} concludes the paper.

%!TEX root = Privacy_Dataset_Paper.tex

\section{Related Work}
\label{sec:related_work}

\subsection{Human Activity Recognition}

Video-based HAR has progressed through several architectural paradigms. Two-stream networks \cite{simonyan2014two} introduced the separation of spatial and temporal processing via RGB and optical flow inputs. 3D convolutional networks, including C3D \cite{tran2015learning}, I3D \cite{carreira2017quo}, and SlowFast \cite{feichtenhofer2019slowfast}, extended this by learning spatiotemporal features directly from video volumes. More recently, transformer-based architectures such as TimeSformer \cite{bertasius2021space} and VideoMAE \cite{tong2022videomae} have achieved state-of-the-art performance by applying self-attention across spatial and temporal dimensions, typically operating on input resolutions of $224 \times 224$ and temporal windows of 16--32 frames.

In parallel, skeleton-based methods using Spatial-Temporal Graph Convolutional Networks (ST-GCNs) \cite{yan2018spatial, xing2023improved} have demonstrated competitive performance by operating on joint coordinate sequences rather than pixel data. These methods are inherently privacy-preserving in that they discard appearance information, but they require reliable pose estimation as a preprocessing step, a dependency that may itself fail under privacy transformations.

\subsection{Visual Privacy in Video}

Visual privacy techniques can be broadly categorized by the type of information they destroy. For a comprehensive survey, see Zhao et al.\ \cite{zhao2025visual}.

\paragraph{Spatial obfuscation.} Blurring \cite{tu2021toward} and pixelation \cite{al2020modeling} reduce spatial resolution within a region of interest, obscuring fine-grained identity features (e.g., facial landmarks) while preserving coarse structure such as body outline and posture. These methods offer weak privacy guarantees, as modern super-resolution networks can partially reconstruct blurred faces \cite{dong2016image}.

\paragraph{Structural abstraction.} Methods such as silhouette extraction, edge detection, and skeleton overlay \cite{gao2024privacy, angelini2019privacy} replace the original appearance with a structural representation. These provide stronger privacy by removing texture and color, but retain body shape and articulation, which may still enable re-identification through gait analysis \cite{nambiar2019gait, elharrouss2021gait}.

\paragraph{Cryptographic transformation.} Pixel-level encryption, including AES-based block scrambling \cite{dumbere2014video, hafsa2022real} and selective encryption of visual features \cite{liu2021video}, aims to render the visual content computationally irrecoverable without the decryption key. The privacy guarantee is tied to cryptographic hardness rather than perceptual degradation. Block-wise scrambling approaches \cite{qi2023privacy, ramasamy2019image} permute spatial blocks using pseudorandom sequences, destroying local structure while preserving global frame statistics. However, the interaction between encrypted pixel data and downstream neural network feature extraction is poorly understood and varies with block size, encryption mode, and spatial granularity.

\paragraph{Learned privacy.} Adversarial and generative approaches train privacy-inducing transformations end-to-end, optimizing for a dual objective: maximizing action recognition accuracy while minimizing identity recognition accuracy \cite{wu2020privacy, wu2018towards, xia2025less}. These methods are architecturally elegant but produce transformations that lack formal privacy guarantees and are specific to the model architecture used during training. Other recent approaches include spatio-temporal privacy mechanisms \cite{li2023stprivacy}, motion difference quantization \cite{kumawat2022privacy}, optical encoding \cite{hinojosa2022privhar}, and differential privacy frameworks \cite{nken2025video}.

\subsection{Existing Datasets and Benchmarks}

% --- TABLE: Comparison of existing privacy-HAR datasets ---
\begin{table}[t]
    \centering
    \caption{Comparison of PrivHAR-Bench with existing datasets used for privacy-preserving HAR evaluation. PrivHAR-Bench is, to our knowledge, the first benchmark providing multiple graduated privacy tiers, background-removed variants, and a standardized evaluation toolkit within a single release.}
    \label{tab:dataset_comparison}
    \small
    \begin{tabular}{lccccccc}
        \toprule
        \textbf{Dataset} & \textbf{Year} & \textbf{Classes} & \textbf{Clips} & \textbf{Privacy Tiers} & \textbf{NoBG} & \textbf{Lossless} & \textbf{Eval Toolkit} \\
        \midrule
        UCF101 \cite{soomro2012ucf101}         & 2012 & 101 & 13,320 & 0 & \texttimes & \texttimes & \texttimes \\
        HMDB51 \cite{kuehne2011hmdb}            & 2011 & 51  & 6,766  & 0 & \texttimes & \texttimes & \texttimes \\
        PA-HMDB51 \cite{wu2020privacy}            & 2020 & 51  & 515  & 1 & \texttimes & \texttimes & \texttimes \\
        Kinetics-400 \cite{kay2017kinetics}     & 2017 & 400 & 306K   & 0 & \texttimes & \texttimes & \texttimes \\
        NTU RGB+D \cite{shahroudy2016ntu}       & 2016 & 60  & 56,880 & 0 & \texttimes & \checkmark & \texttimes \\
        NTU RGB+D 120 \cite{liu2019ntu}         & 2019 & 120 & 114,480 & 0 & \texttimes & \checkmark & \texttimes \\
        %PA-HMDB \cite{wu2020privacy}            & 2020 & 51  & 515    & 1  & \texttimes & \texttimes & \texttimes \\
        Bullying10K \cite{dong2023bullying10k}   & 2023 & 10  & 10,000 & 0  & \texttimes & \texttimes & \texttimes \\
        \midrule
        \textbf{PrivHAR-Bench (Ours)}   & 2026 & 15 & 1,932 & 9 & \checkmark & \checkmark & \checkmark \\
        \bottomrule
    \end{tabular}
\end{table}

The dominant benchmarks for HAR: UCF101 \cite{soomro2012ucf101}, HMDB51 \cite{kuehne2011hmdb}, Kinetics \cite{kay2017kinetics}, and NTU RGB+D \cite{shahroudy2016ntu}, were designed to evaluate recognition accuracy, not privacy-utility trade-offs. They provide clear, unmodified video and do not include privacy-transformed variants.

Wu et al.\ \cite{wu2020privacy} introduced PA-HMDB, a privacy-annotated subset of HMDB51, and proposed an adversarial framework for privacy-preserving action recognition. While PA-HMDB provides privacy-relevant annotations (action labels and privacy attributes), it offers only a single learned transformation rather than a graduated spectrum. Dong et al.\ \cite{dong2023bullying10k} released Bullying10K, a neuromorphic dataset that achieves privacy through the event camera modality itself, but this is hardware-specific and not applicable to standard RGB pipelines. Ishikawa et al.\ \cite{ishikawa2024learnable} proposed learnable cube-based video encryption but evaluated on standard benchmarks without releasing a dedicated privacy benchmark dataset.

Studies on privacy-preserving HAR typically construct their own evaluation sets ad hoc: selecting a subset of classes from an existing benchmark, applying their proposed transformation, and reporting results on non-standardized splits \cite{li2023stprivacy, kumawat2022privacy, zhang2025stealthguard}. This makes cross-paper comparison unreliable, as differences in class selection, split strategy, input resolution, and temporal sampling all confound the comparison.

To our knowledge, no publicly available benchmark provides (a) multiple graduated privacy tiers applied to the same source videos, (b) background-removed variants for context bias control, and (c) a standardized evaluation protocol with a released toolkit. PrivHAR-Bench is designed to fill this gap.

%!TEX root = Privacy_Dataset_Paper.tex

\section{Benchmark Design}
\label{sec:benchmark_design}

PrivHAR-Bench is designed around three principles: \textit{graduated privacy}, \textit{context isolation}, and \textit{evaluation reproducibility}. This section describes the design decisions and their rationale.

\subsection{The Privacy Spectrum}
\label{subsec:privacy_spectrum}

A binary comparison between clear and encrypted video reveals only whether a method tolerates a specific transformation, not how it degrades across a continuum of privacy strength. PrivHAR-Bench organizes privacy transformations into three tiers of increasing strength, each targeting a different category of visual information:

\paragraph{Tier 1: Spatial Obfuscation (Gaussian Blur).} A Gaussian blur with kernel standard deviation $\sigma = 15$ is applied to all pixels within the detected Region of Interest (ROI). This preserves gross spatial structure: body outline, posture, limb position, while destroying fine-grained features such as facial landmarks, skin texture, and clothing detail. Tier 1 serves as the ``easy'' privacy condition: if a recognition model fails at this level, it indicates a fundamental architectural limitation rather than a privacy-induced challenge.

\paragraph{Tier 2: Structural Abstraction (Edge Extraction).} Canny edge detection is applied to the ROI, with all non-edge pixels within the ROI set to black. This removes all texture and color information, retaining only the structural contours of the human body. Tier 2 tests whether movement dynamics encoded in body contour trajectories are sufficient for recognition in the absence of appearance features.

\paragraph{Tier 3: Cryptographic Block Permutation (AES Block Scrambling).} The ROI is divided into non-overlapping square blocks of size $B \times B$ pixels, and these blocks are permuted according to a pseudorandom sequence derived from an AES-CTR keystream. This destroys both spatial structure and texture within the ROI, preserving only the gross temporal dynamics of pixel intensity changes across frames. Because the privacy-utility trade-off is directly controlled by block size $B$, Tier 3 is provided at three granularities:

\begin{itemize}
 	\item \textbf{Tier 3a ($B = 16$):} Coarse scrambling. Preserves more local structure within each block.
 	 \item \textbf{Tier 3b ($B = 8$):} Medium granularity.
    \item \textbf{Tier 3c ($B = 4$):} Fine-grained scrambling. Maximizes spatial destruction.
\end{itemize}

The choice of block sizes $B \in \{4, 8, 16\}$ is empirically justified in Section \ref{sec:experiments} through a face verification experiment measuring the block size at which a state-of-the-art face recognition model (ArcFace \cite{deng2019arcface}) can no longer detect faces in the scrambled ROI.

% --- FIGURE: Privacy Spectrum Diagram ---
\begin{figure}[t]
    \centering
    \includegraphics[width=\textwidth]{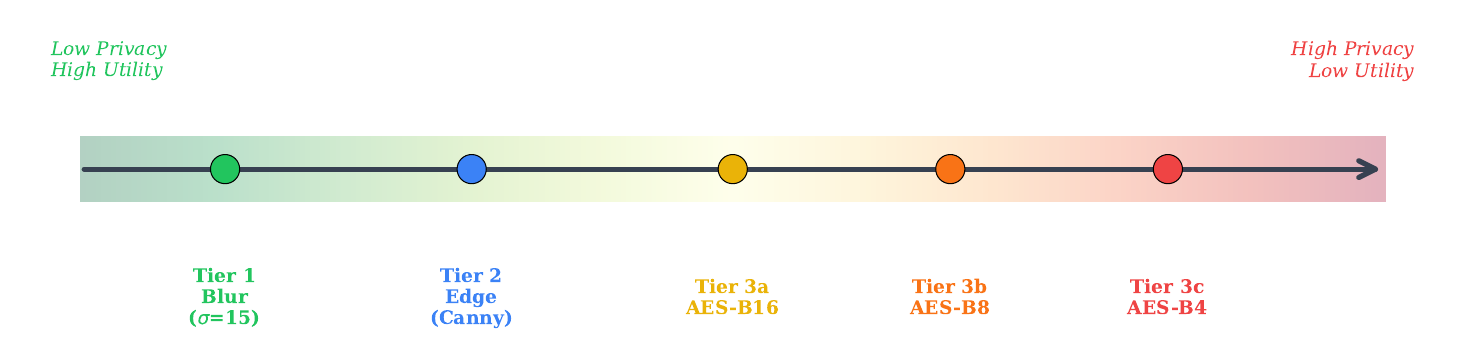}
    \caption{The PrivHAR-Bench privacy spectrum. Each tier progressively destroys a different category of visual information, from fine-grained appearance (Tier 1) through structural contour (Tier 2) to spatial pixel arrangement (Tier 3). Tier 3 is further parameterized by block size $B$.}
    \label{fig:privacy_spectrum}
\end{figure}

\subsection{Context Bias Control}
\label{subsec:context_bias}

HAR models are known to exploit background context for classification \cite{rastegar2024background, zhou2025seeing, fukuzawa2025can}. A model may correctly classify ``Tai Chi'' not by recognizing the movement but by recognizing the park or gym in which it is performed. When privacy transformations are applied only to the human ROI, the background remains fully visible, creating a shortcut that bypasses the privacy transformation entirely.

To control for this confound, PrivHAR-Bench provides a \textit{NoBG} (No Background) variant for every Tier 3 video. In NoBG variants, all pixels outside the detected ROI are set to zero (black), leaving only the scrambled human region visible. Any model evaluated on the NoBG variant must derive its classification signal exclusively from the transformed ROI.

% --- FIGURE: NoBG comparison ---
\begin{figure}[t]
    \centering
    \includegraphics[width=\textwidth]{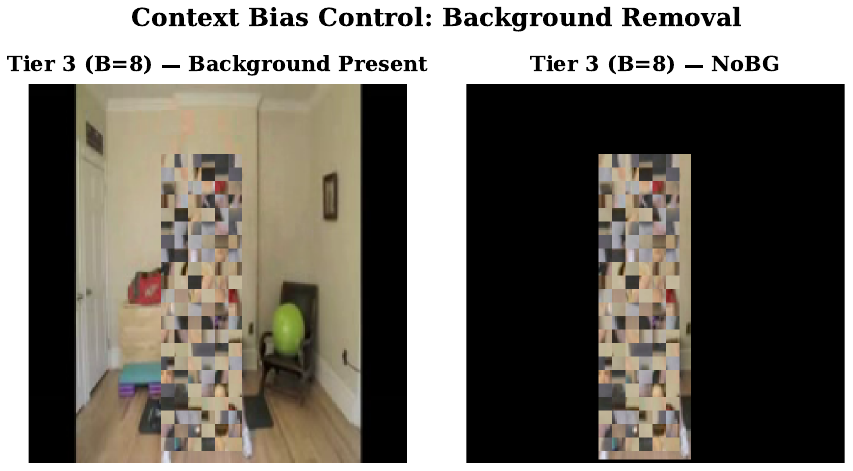}
    \caption{Illustration of context bias control. Left: Tier 3 ($B=8$) with background preserved, a model could exploit environmental cues. Right: the NoBG variant of the same frame, isolating the scrambled human region.}
    \label{fig:nobg_comparison}
\end{figure}

\subsection{Class Selection}
\label{subsec:class_selection}

PrivHAR-Bench is constructed from a curated subset of UCF101 \cite{soomro2012ucf101}. Classes were selected according to three criteria:

\begin{enumerate}
    \item \textbf{Articulation diversity.} The selected classes must collectively cover a range of human motion patterns: fine upper-body manipulation (e.g., BrushingTeeth), full-body cyclic motion (e.g., JumpingJack), explosive multi-joint movement (e.g., CleanAndJerk), locomotion (e.g., WalkingWithDog), and standing task-oriented motion (e.g., WritingOnBoard).
    \item \textbf{Minimal object/context dependency.} Classes where classification relies primarily on a visible object (e.g., Biking, HorseRiding) or a distinctive environment (e.g., Surfing, IceDancing) were excluded, as these confounds undermine the purpose of a privacy benchmark.
    \item \textbf{Detection feasibility.} Each candidate class was validated by running YOLOv8n-Pose \cite{jocher2023yolov8, dong2024enhanced} on a sample of clips. Classes where per-frame person detection fell below 70\% for a substantial portion of clips (e.g., BandMarching, PullUps, Rowing, HandstandWalking) were excluded due to unreliable ROI extraction.
\end{enumerate}

% --- TABLE: Final class list ---
\begin{table}[t]
    \centering
    \caption{The 15 selected classes, grouped by motion category. Clip count and group count are from the UCF101 source. Detection rate is the median per-frame person detection rate across all clips in the class using YOLOv8n-Pose.}
    \label{tab:class_list}
    \small
    \begin{tabular}{llccc}
        \toprule
        \textbf{Category} & \textbf{Class} & \textbf{Clips} & \textbf{Groups} & \textbf{Med. Det. Rate} \\
        \midrule
        \multirow{6}{*}{ADL / Care}
            & BrushingTeeth    & 131 & 25 & 100.0\% \\
            & Haircut          & 130 & 25 & 99.5\% \\
            & MoppingFloor     & 110 & 25 & 99.7\% \\
            & ApplyEyeMakeup   & 145 & 25 & 100.0\% \\
            & BabyCrawling     & 132 & 25 & 97.5\% \\
            & ShavingBeard     & 161 & 25 & 100.0\% \\
        \midrule
        \multirow{8}{*}{Full-Body Articulation}
            & BodyWeightSquats & 112 & 25 & 100.0\% \\
            & Lunges           & 127 & 25 & 99.2\% \\
            & TaiChi           & 100 & 25 & 100.0\% \\
            & JumpRope         & 144 & 25 & 100.0\% \\
            & WritingOnBoard   & 152 & 25 & 100.0\% \\
            & WallPushups      & 130 & 25 & 100.0\% \\
            & JumpingJack      & 123 & 25 & 100.0\% \\
            & CleanAndJerk     & 112 & 25 & 99.6\% \\
        \midrule
        Tracking
            & WalkingWithDog   & 123 & 25 & 80.0\% \\
        \midrule
        \multicolumn{2}{l}{\textbf{Total}} & \textbf{1,932} & & \\
        \bottomrule
    \end{tabular}
\end{table}

\subsection{Temporal Standardization}

All videos are clipped to a standardized window of 32 frames, taken from the temporal center of each source clip. This design choice serves two purposes. First, it ensures compatibility with the dominant architectures: SlowFast \cite{feichtenhofer2019slowfast} typically operates on 32--64 frame inputs, while TimeSformer \cite{bertasius2021space} and VideoMAE \cite{tong2022videomae} commonly use 16--32 frames. Users requiring 16-frame inputs can trivially subsample by stride-2 or slicing. Second, a bounded temporal window limits the temporal information available for potential re-identification attacks based on gait periodicity, which typically require multiple complete gait cycles \cite{nambiar2019gait}.

We note that this temporal constraint \textit{reduces} but does not \textit{eliminate} the risk of temporal re-identification. Research on gait recognition has demonstrated successful identification from short sequences under controlled conditions \cite{chao2019gaitset, fan2020gaitpart}. We discuss this limitation in Section \ref{sec:limitations}.

\subsection{Lossless Storage Format}

Lossy video codecs (H.264, H.265) apply spatial quantization and chroma subsampling that smooths high-frequency content. AES-scrambled pixel blocks constitute high-frequency pseudorandom noise, which lossy codecs will partially destroy during encoding. The mathematical properties of the scrambling, and consequently, any privacy metrics computed on the data, would be corrupted before the researcher even loads the file.

PrivHAR-Bench distributes all frames as lossless PNG image sequences. This ensures that the pixel values a researcher loads are byte-identical to the values produced by the generation pipeline. The storage cost of lossless encoding for pseudorandom data is substantial (23.0 GB total; see Section \ref{sec:dataset_generation}), but it is the only format that preserves the integrity of the privacy transformations.

%!TEX root = Privacy_Dataset_Paper.tex

\section{Dataset Generation Pipeline}
\label{sec:dataset_generation}

The PrivHAR-Bench generation pipeline transforms raw UCF101 videos into multi-tier lossless frame sequences through a deterministic, reproducible process. The complete pipeline is released as open-source code with pinned dependencies and fixed random seeds.

\subsection{Source Data}

The source material is the UCF101 Action Recognition Dataset \cite{soomro2012ucf101}, consisting of 13,320 clips across 101 action classes at a native resolution of $320 \times 240$ pixels. From this, we select 15 classes comprising 1,932 total clips (see Table~\ref{tab:class_list}). We acknowledge that UCF101's native resolution of $320 \times 240$ is substantially lower than modern surveillance systems. We address the implications of this limitation in Section~\ref{sec:limitations}.

\subsection{ROI Detection}
\label{subsec:roi_detection}

Person detection and pose estimation are performed using YOLOv8n-Pose \cite{jocher2023yolov8, dong2024enhanced}. When multiple persons are detected in a frame, the detection with the largest bounding box area is selected as the primary subject. Detections with confidence below 0.5 are discarded.

For each frame, the pipeline outputs:
\begin{itemize}
    \item A bounding box $[x_{\min}, y_{\min}, x_{\max}, y_{\max}]$ defining the ROI.
    \item A binary mask $M$ where $M_{ij} = 1$ for pixels inside the ROI and $M_{ij} = 0$ otherwise.
    \item 17 COCO-format keypoints, each with coordinates $(x, y)$ and a detection confidence $c \in [0, 1]$.
\end{itemize}

Frames in which no person is detected with sufficient confidence are retained in the dataset with a null annotation. The per-video detection rate is recorded in the metadata (see Section~\ref{subsec:metadata}). Across all 1,932 clips, the mean per-video detection rate is 94.7\%, with 13 of 15 classes exceeding 99\% median detection. The lowest-performing class is WalkingWithDog (80.0\% median), attributable to the small apparent size of subjects in wide-angle outdoor shots.

% --- FIGURE: Pipeline overview ---
\begin{figure}[t]
    \centering
    \includegraphics[width=\textwidth]{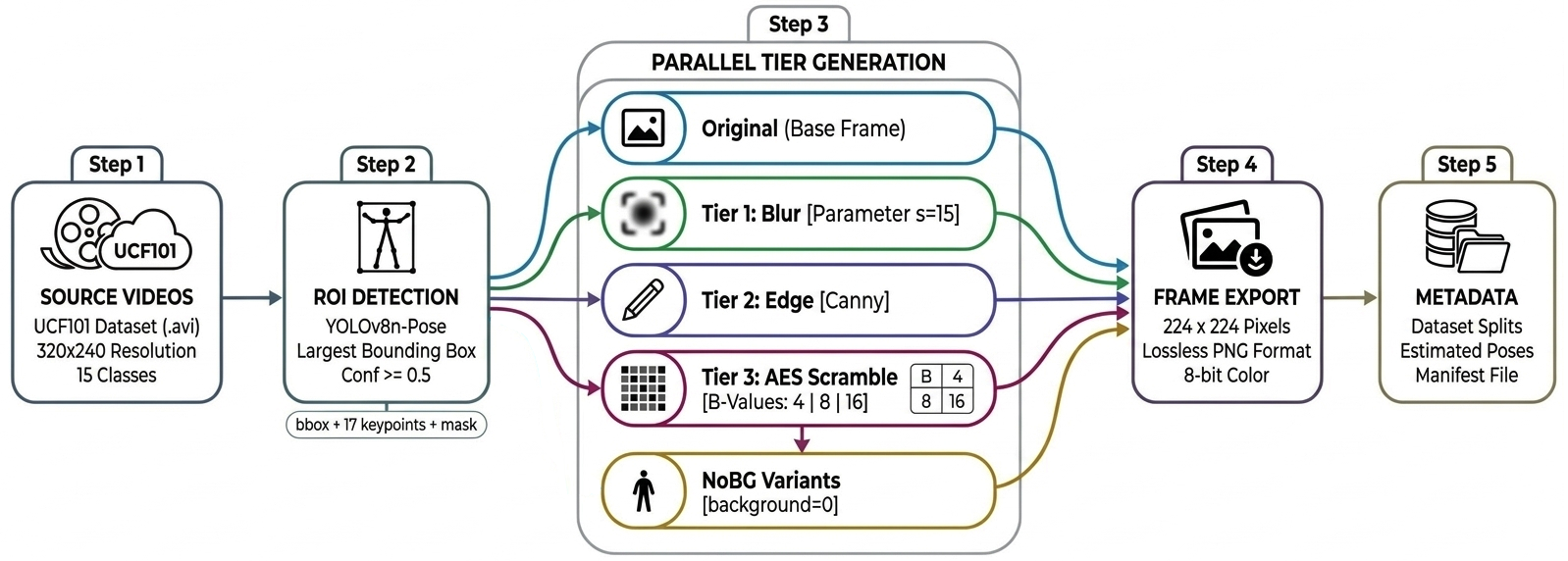}
    \caption{Overview of the PrivHAR-Bench generation pipeline. Each source video passes through a shared ROI detection stage, after which all privacy tiers are generated in parallel using the same bounding box and mask.}
    \label{fig:pipeline_overview}
\end{figure}

\subsection{Tier Generation}

All tiers operate on the same per-frame ROI bounding box and mask, ensuring spatial consistency across variants.

\paragraph{Tier 1: Gaussian Blur.} A Gaussian kernel with $\sigma = 15$ is applied to the ROI region. Formally, for pixel $p$ inside the ROI:
\begin{equation}
    p'_{\text{blur}} = (G_\sigma * I)(p), \quad \forall p \in \text{ROI}
\end{equation}
where $G_\sigma$ is a 2D Gaussian kernel and $I$ is the input frame. Pixels outside the ROI are unchanged.

\paragraph{Tier 2: Canny Edge.} Canny edge detection is applied to the ROI with thresholds $t_{\text{low}} = 50$ and $t_{\text{high}} = 150$. The output frame contains edge pixels (white) on a black background within the ROI. All pixels outside the ROI are set to black:
\begin{equation}
    p'_{\text{edge}} = \begin{cases}
        255 & \text{if } p \in \text{ROI} \text{ and } \text{Canny}(I, t_{\text{low}}, t_{\text{high}})(p) = 1 \\
        0   & \text{otherwise}
    \end{cases}
\end{equation}

\paragraph{Tier 3: AES Block Permutation.} The ROI is divided into non-overlapping blocks of size $B \times B$ pixels. A pseudorandom permutation of the block indices is generated using a keystream derived from AES-128 in Counter (CTR) mode. A single global key $K$ is fixed across the dataset. The AES-CTR nonce is derived by computing SHA-256 over the concatenation of the video identifier, frame index, and block size, and truncating to 8 bytes. This produces a unique permutation for every combination of video, frame, and block size. The blocks are then spatially rearranged according to this permutation:
\begin{equation}
	\{b_1, b_2, \ldots, b_n\} \xrightarrow{\pi_K} \{b_{\pi(1)}, b_{\pi(2)}, \ldots, b_{\pi(n)}\}
\end{equation}
where $\pi_K$ is the AES-derived permutation and $b_i$ denotes the $i$-th block. The permutation $\pi_K$ is recomputed independently for each frame: the nonce incorporates the frame index, so no two frames within a clip share the same block arrangement. This eliminates temporal block correspondence; a model cannot track the spatial trajectory of a specific block across frames. Pixels within each block retain their original values; only the spatial arrangement of blocks is altered. Pixels outside the ROI are unchanged. This process is applied independently for block sizes $B \in \{4, 8, 16\}$, yielding three sub-tiers (3a, 3b, 3c). 

\paragraph{NoBG Variants.} For each Tier~3 variant, an additional NoBG version is produced by applying the pre-computed mask:
\begin{equation}
    p'_{\text{NoBG}} = p'_{\text{tier3}} \cdot M(p)
\end{equation}
where $M(p) \in \{0, 1\}$ is the binary ROI mask. This zeroes all background pixels.

\subsection{Frame Export and Resolution}

All output frames are resized to $224 \times 224$ pixels and saved as lossless 8-bit PNG files. The resize is performed on the \textit{source frame} prior to tier transformation to ensure consistent spatial dimensions across all tiers. We use bilinear interpolation for resizing.

\subsection{Metadata and Annotations}
\label{subsec:metadata}

A structured \texttt{annotations.json} file records per-video metadata:

\begin{verbatim}
{
  "video_id": "00001",
  "source_file": "v_BrushingTeeth_g01_c01.avi",
  "class": "BrushingTeeth",
  "split": "train",
  "source_fps": 25,
  "total_frames": 120,
  "clip_frames": 32,
  "detection_rate": 0.98,
  "roi_bbox_mean": [45, 30, 180, 220]
}
\end{verbatim}

\subsection{Train/Test Splits}

Splits are performed by UCF101 group identifier to prevent identity leakage: all clips from the same group appear exclusively in either the training or test set. Groups 1--19 are assigned to training and groups 20--25 to testing, yielding a 75\%/25\% split: 1,452 training clips and 480 test clips. The split files (\texttt{train\_split.txt}, \texttt{test\_split.txt}) are provided as fixed lists of video IDs and must not be regenerated by users.

\subsection{Reproducibility}

The generation pipeline enforces deterministic execution:
\begin{itemize}
    \item All random seeds are fixed: \texttt{torch.manual\_seed(42)}, \texttt{numpy.random.seed(42)}, \texttt{random.seed(42)}.
    \item PyTorch deterministic mode is enabled: \texttt{torch.use\_deterministic\_algorithms(True)}.
    \item cuDNN benchmarking is disabled: \texttt{torch.backends.cudnn.benchmark = False}.
    \item The YOLOv8n-Pose weight file is verified by SHA-256 hash at runtime. A hash mismatch produces a fatal error.
    \item All library versions are pinned in \texttt{requirements.txt}.
\end{itemize}

Non-determinism arising from CUDA atomic operations on different GPU architectures may cause minor per-pixel variations in pose estimation outputs. These variations do not affect the bounding box (which is rounded to integer coordinates) or the encryption (which operates on the bounding box, not keypoints). The AES block permutation (Section 4.3) requires \texttt{pycryptodome} for canonical output. The released pipeline includes a SHA-256-based CSPRNG fallback for environments where \texttt{pycryptodome} is unavailable; however, the two paths use different pseudorandom generators and produce different permutations for the same input. The distributed dataset was generated exclusively using the AES-CTR path. Users must install \texttt{pycryptodome} to reproduce byte-identical outputs; the fallback is provided only for pipeline inspection and is not suitable for dataset regeneration.

\subsection{Dataset Statistics}

Table~\ref{tab:dataset_stats} summarizes the dataset composition.

% --- TABLE: Dataset size and storage ---
\begin{table}[t]
    \centering
    \caption{PrivHAR-Bench dataset composition and storage requirements.}
    \label{tab:dataset_stats}
    \small
    \begin{tabular}{lc}
        \toprule
        \textbf{Property} & \textbf{Value} \\
        \midrule
        Source dataset & UCF101 \cite{soomro2012ucf101} \\
        Number of classes & 15 \\
        Total source clips & 1,932 \\
        Frames per clip & 32 \\
        Frame resolution & $224 \times 224$ \\
        Number of tiers & 9 (Original + Blur + Edge + B4 + B8 + B16 + 3$\times$NoBG) \\
        Total frame sequences & 17,388 (1,932 clips $\times$ 9 tiers) \\
        Storage format & Lossless PNG \\
        Total dataset size & 23.0 GB \\
        Train / Test split & 1,452 / 480 (group-based, 75\%/25\%) \\
        Distribution & Zenodo (segmented ZIP archives) \\
        \bottomrule
    \end{tabular}
\end{table}

\subsection{Directory Structure}

\begin{verbatim}
/PrivHAR-Bench_v1.0.0
|-- /Original
|-- /Tier1_Blur
|-- /Tier2_Edge
|-- /Tier3_AES_B4
|-- /Tier3_AES_B8
|-- /Tier3_AES_B16
|-- /Tier3_AES_B4_NoBG
|-- /Tier3_AES_B8_NoBG
|-- /Tier3_AES_B16_NoBG
|-- /Estimated_Poses        (17 keypoints per frame, JSON)
|-- annotations.json
|-- train_split.txt
|-- test_split.txt
|-- manifest.json           (SHA-256 per file)
\-- CHANGELOG.md
\end{verbatim}

%!TEX root = Privacy_Dataset_Paper.tex

\section{Evaluation Protocol}
\label{sec:evaluation_protocol}

PrivHAR-Bench prescribes a fixed evaluation protocol to ensure cross-study comparability. All metrics, split definitions, and input specifications are implemented in the released \texttt{eval.py} toolkit.

\subsection{Recognition Metrics}

\paragraph{Top-1 Accuracy.} The primary recognition metric. A model is trained (or fine-tuned) on the training split of a given tier and evaluated on the corresponding test split. Accuracy is reported per-tier, enabling direct comparison across privacy levels.

\paragraph{Cross-Tier Accuracy Drop ($\Delta_{\text{acc}}$).} To quantify the privacy-utility trade-off, we define the accuracy drop relative to the Original (clear) tier:
\begin{equation}
    \Delta_{\text{acc}}(T) = \text{Acc}_{\text{Original}} - \text{Acc}_{T}
\end{equation}
where $T$ denotes a specific tier. A method that maintains small $\Delta_{\text{acc}}$ across tiers demonstrates robustness to privacy transformations.

\subsection{Privacy Metrics}

\paragraph{Structural Similarity Index (SSIM).} SSIM \cite{wang2004image} is computed between each transformed frame and its original counterpart within the ROI. SSIM captures perceptual similarity; lower SSIM indicates greater visual privacy. We report SSIM exclusively within the ROI bounding box to avoid inflating the metric with unchanged background pixels. For a discussion of SSIM versus alternative image quality metrics, see Hore and Ziou \cite{hore2010image}.

\paragraph{Peak Signal-to-Noise Ratio (PSNR).} Computed within the ROI, providing a pixel-level distortion measure \cite{sara2019image}. Like SSIM, lower PSNR corresponds to stronger privacy.

\paragraph{Face Detection Failure Rate.} For clips containing detectable faces, we apply ArcFace \cite{deng2019arcface} to both the original and transformed frames. The \textit{failure rate} is the proportion of originally-detectable faces that become undetectable after transformation, i.e., ArcFace cannot locate a face in the scrambled ROI at all. This is a stricter measure than embedding similarity: the face is not merely dissimilar, but entirely absent from the detector's perspective. Sun et al.\ \cite{sun2023privacy} demonstrate that standard image quality metrics may not faithfully reflect human privacy perception; we include face detection failure as a task-specific complement to SSIM/PSNR.

\subsection{Composite Privacy-Utility Score}

We define a composite Privacy-Utility ($PU$) score:
\begin{equation}
    PU = \frac{\text{Acc}_{T}}{\text{Acc}_{\text{Original}}} \times (1 - \text{SSIM}_{T})
\end{equation}
A high $PU$ score indicates that a model maintains recognition accuracy (high first term) on data with strong visual privacy (low SSIM, yielding high second term). This single scalar enables ranking of methods across the privacy-utility space, complementing the multi-dimensional analysis advocated by Abbasi et al.\ \cite{abbasi2024trading}.

\subsection{Prescribed Evaluation Configurations}

To ensure comparability, we require that all evaluations using PrivHAR-Bench report results under the following configurations:

% --- TABLE: Required evaluation configurations ---
\begin{table}[t]
    \centering
    \caption{Required evaluation configurations for PrivHAR-Bench benchmark submissions. ``Train Tier'' indicates the privacy level of training data; ``Eval Tier'' indicates the privacy level of test data. Config A measures within-tier performance; Config B measures generalization from clear training data to encrypted test data.}
    \label{tab:eval_configs}
    \small
    \begin{tabular}{clll}
        \toprule
        \textbf{Config} & \textbf{Train Tier} & \textbf{Eval Tier} & \textbf{Purpose} \\
        \midrule
        A & Same as Eval & Each tier & Within-tier accuracy \\
        B & Original (Clear) & Each tier & Cross-domain generalization \\
        C & Tier3\_B8\_NoBG & Tier3\_B8\_NoBG & Context-free recognition \\
        \bottomrule
    \end{tabular}
\end{table}

\subsection{Baseline Model}

We provide baseline results using R3D-18 \cite{tran2018closer}, a 3D ResNet-18 architecture pre-trained on Kinetics-400 and fine-tuned on PrivHAR-Bench. R3D-18 processes 32-frame clips at $224 \times 224$ resolution, directly matching the PrivHAR-Bench format. The model is trained with Adam optimization ($\text{lr} = 3 \times 10^{-4}$, weight decay $10^{-4}$) and cosine annealing. The Original tier was trained for 15 epochs; all privacy tiers were trained for 5 epochs. This asymmetry slightly disadvantages the privacy tiers and is noted as a limitation. All hyperparameters, checkpoints, and training logs are released with the benchmark.

\subsection{Evaluation Toolkit}

The released \texttt{eval.py} script accepts:
\begin{itemize}
    \item A directory of model predictions (one CSV file per tier, mapping video IDs to predicted labels).
    \item The PrivHAR-Bench \texttt{annotations.json} and test split file.
\end{itemize}

It outputs: per-class Top-1 accuracy, overall Top-1 accuracy, ROI-SSIM, ROI-PSNR, Face Detection Failure Rate, $PU$ score, and $\Delta_{\text{acc}}$ for each tier. All metrics are computed identically for all users, eliminating implementation-dependent variation.

%!TEX root = Privacy_Dataset_Paper.tex

\section{Experiments}
\label{sec:experiments}

This section presents baseline results on PrivHAR-Bench, empirical justification for key design parameters, and a privacy audit assessing the dataset's resilience to identity recovery.

\subsection{Baseline Recognition Results}

% --- TABLE: Main results table (Config A) ---
\begin{table}[t]
    \centering
    \caption{Top-1 accuracy (\%) of the R3D-18 baseline across PrivHAR-Bench privacy tiers under Config A (train and test on the same tier). Accuracy generally decreases with increasing privacy strength. The NoBG variant isolates the contribution of the scrambled human region from background context.}
    \label{tab:main_results}
    \small
    \begin{tabular}{lccccccc}
        \toprule
        \textbf{Model} & \textbf{Original} & \textbf{Blur} & \textbf{Edge} & \textbf{B16} & \textbf{B8} & \textbf{B4}  & \textbf{B8-NoBG} \\
        \midrule
        R3D-18  & 88.8 & 66.5 & 66.2 & 66.2  & 64.4 &  63.1 & 53.5 \\
        \midrule
        $\Delta_{\text{acc}}$ & --- & 22.3 & 22.6 & 22.6  & 24.4 &  25.7 & 35.3 \\
        \bottomrule
    \end{tabular}
\end{table}

% --- TABLE: Config B results (cross-domain) ---
\begin{table}[t]
    \centering
    \caption{Top-1 accuracy (\%) under Config B: model trained on Original (clear) data and evaluated on each privacy-transformed tier without fine-tuning. This measures out-of-distribution robustness. Random chance for 15 classes is 6.7\%.}
    \label{tab:cross_domain_results}
    \small
    \begin{tabular}{lcccccc}
        \toprule
        \textbf{Model} & \textbf{Blur} & \textbf{Edge} & \textbf{B16}& \textbf{B8} &   \textbf{B4}  & \textbf{B8-NoBG} \\
        \midrule
        R3D-18  & 48.1 & 6.9 & 39.2  & 34.4 & 35.6 & 4.8 \\
        \bottomrule
    \end{tabular}
\end{table}

Table~\ref{tab:main_results} shows Config~A results. The R3D-18 baseline achieves 88.8\% on Original (clear) video, declining to 63.1--66.5\% across privacy tiers. Tier~1 (Blur), Tier~2 (Edge), and Tier~3a ($B=16$) all converge at approximately 66\%, suggesting that with sufficient training data, R3D-18 can partially adapt to each transformation type. Finer-grained scrambling ($B=4$) produces the largest accuracy drop among the Tier~3 variants, consistent with greater spatial destruction. The B8-NoBG column corresponds to Config C (Table 4): the model trained and evaluated exclusively on the background-removed variant achieves 53.5\%, confirming that removing environmental context imposes an additional 10.9 percentage point penalty beyond spatial scrambling alone. We note that the per-frame permutation scheme (Section 4.3) eliminates temporal block correspondence, preventing models from tracking individual blocks across frames. This strengthens the privacy guarantee but also makes the recognition task strictly harder than a fixed-per-video permutation would; the Tier 3 accuracy values reported here should be interpreted in this context.

Table~\ref{tab:cross_domain_results} shows Config~B results. Without fine-tuning, accuracy on Edge drops to 6.9\%, at random chance, demonstrating that features learned from RGB video do not transfer to edge representations. AES-scrambled tiers retain 34--39\% accuracy, indicating that some low-level features (e.g., color statistics, motion energy) survive the block permutation. The B8-NoBG result of 4.8\% (below random chance) confirms that the Original-trained model has learned essentially nothing transferable when both background and spatial structure are removed.

% --- FIGURE: Accuracy vs. Tier plot ---
\begin{figure}[t]
    \centering
    \includegraphics[width=\textwidth]{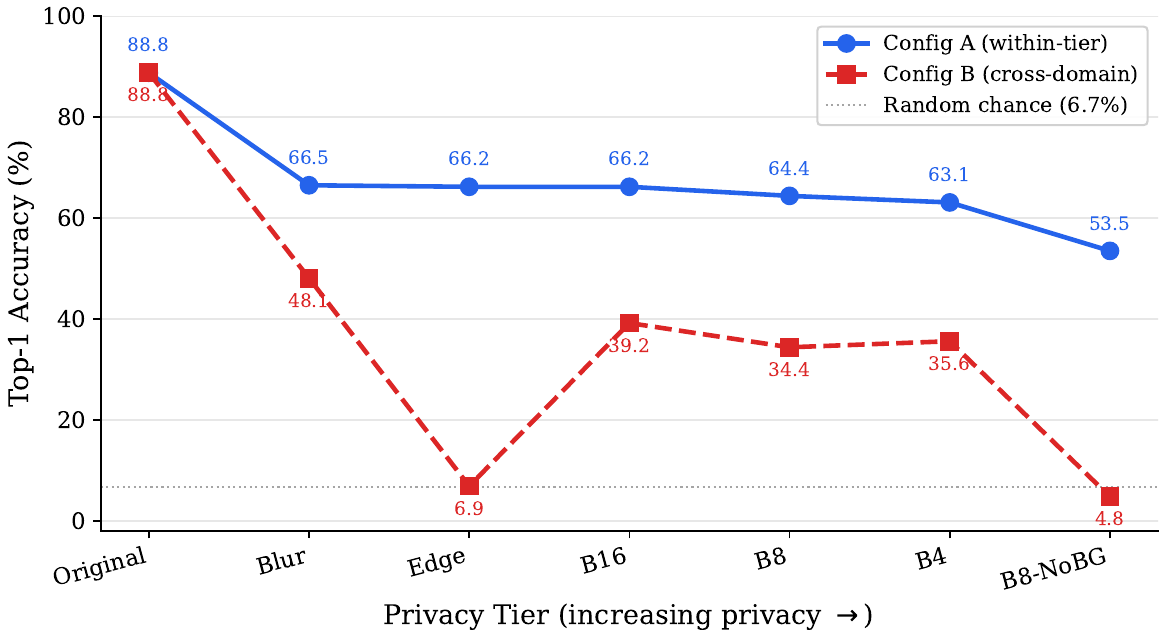}
    %\missing{FIGURE 5: Line plot. X-axis: privacy tiers ordered by increasing privacy (Original, Blur, B16, B8, B4, B8-NoBG). Y-axis: Top-1 accuracy (\%). Solid line: Config A. Dashed line: Config B. Data points: Original: 88.8/88.8, Blur: 66.5/48.1, B16: 66.2/39.2, B8: 64.4/34.4, B4: 63.1/35.6, B8-NoBG: 53.5/4.8.}
	\caption{Top-1 accuracy as a function of privacy tier for the R3D-18 baseline. Solid line: Config A (within-tier training). Dashed line: Config B (cross-domain, trained on Original only). The divergence between curves quantifies the domain gap introduced by each transformation. \textit{Note:} The B8-NoBG condition combines two independent manipulations: cryptographic block scrambling (privacy) and background removal (context bias control). It is plotted at the rightmost position for visual continuity, but it is not strictly a point on the same linear privacy scale as the preceding tiers; it additionally removes an orthogonal confound (see Section 3.2 and Figure 3).}
    \label{fig:accuracy_vs_tier}
\end{figure}

\subsection{Context Bias Analysis}

To quantify the effect of background context, we compare model accuracy on Tier~3 ($B=8$) with and without the background present.

% --- TABLE: Context bias results ---
\begin{table}[t]
    \centering
    \caption{Effect of background removal on Tier 3 ($B=8$) recognition accuracy. The 10.9 percentage point gap confirms that the model derives substantial classification signal from environmental context \cite{rastegar2024background, zhou2025seeing}.}
    \label{tab:context_bias}
    \small
    \begin{tabular}{lccc}
        \toprule
        \textbf{Model} & \textbf{Tier3-B8 (With BG)} & \textbf{Tier3-B8 (NoBG)} & \textbf{$\Delta$} \\
        \midrule
        R3D-18 (Config A)  & 64.4\% & 53.5\% & $-$10.9 pp \\
        R3D-18 (Config B)  & 34.4\% & 4.8\%  & $-$29.6 pp \\
        \bottomrule
    \end{tabular}
\end{table}

Under Config~A, removing the background reduces accuracy by 10.9 percentage points (64.4\% $\to$ 53.5\%). This means that approximately 17\% of the model's apparent performance on scrambled video was attributable to background cues rather than recognition of the transformed human region. Under Config~B, the effect is even more dramatic: background removal reduces accuracy from 34.4\% to 4.8\%, a 29.6 percentage point drop. This confirms that the NoBG variant is a necessary component of the benchmark for rigorous evaluation. Baseline results are reported for B8-NoBG only. B = 8 was selected as the representative NoBG condition because it occupies the midpoint of the block size range; B4-NoBG and B16-NoBG are distributed in the dataset and available for evaluation but are omitted here to avoid redundancy. Users requiring comprehensive NoBG coverage across all block sizes can evaluate these tiers using the released toolkit.

\subsection{Empirical Block Size Justification}
\label{subsec:block_size_experiment}

We assess the effect of block scrambling on face identity using a pre-trained ArcFace \cite{deng2019arcface} model (buffalo\_l, ResNet-50~\cite{he2016deep} backbone).

\paragraph{Protocol.} We sample 500 frames containing detectable faces from the Original videos. For each face, we apply block scrambling at $B \in \{2, 4, 8, 16, 32\}$ and attempt face detection on the scrambled frame using the same ArcFace pipeline.

\paragraph{Results.} The primary finding is that block scrambling at all tested granularities \textit{prevents face detection} in the vast majority of cases. Of 500 original faces, only 53--69 (10.6--13.8\%) remained detectable post-scrambling, with no meaningful trend across block sizes (Table~\ref{tab:arcface_results}). The scrambling destroys facial structure below the face detector's threshold rather than gradually degrading embedding similarity. For the minority of faces that remained detectable, typically large, frontal faces where the ROI bounding box extends well beyond the face region, the conditional cosine similarity remained high (median 0.91--0.97), indicating that some identity information survives when the face is still localizable.

% --- TABLE: ArcFace results ---
\begin{table}[t]
    \centering
    \caption{ArcFace face detection and identity verification results after block scrambling. ``Det.'' is the number of faces (out of 500) where ArcFace could detect a face post-scrambling. ``Fail Rate'' is the percentage of faces rendered undetectable. Conditional similarity is computed only over the detected subset.}
    \label{tab:arcface_results}
    \small
    \begin{tabular}{cccccc}
        \toprule
        \textbf{Block Size $B$} & \textbf{Det.} & \textbf{Fail Rate} & \textbf{Mean Sim.} & \textbf{Median Sim.} & \textbf{$<$0.2} \\
        \midrule
        2  & 56 & 88.8\% & 0.631 & 0.912 & 27\% \\
        4  & 54 & 89.2\% & 0.660 & 0.931 & 24\% \\
        8  & 53 & 89.4\% & 0.658 & 0.955 & 24\% \\
        16 & 56 & 88.8\% & 0.684 & 0.965 & 23\% \\
        32 & 69 & 86.2\% & 0.670 & 0.967 & 22\% \\
        \bottomrule
    \end{tabular}
\end{table}

% --- FIGURE: ArcFace box plot ---
\begin{figure}[t]
    \centering
    \includegraphics[width=0.8\textwidth]{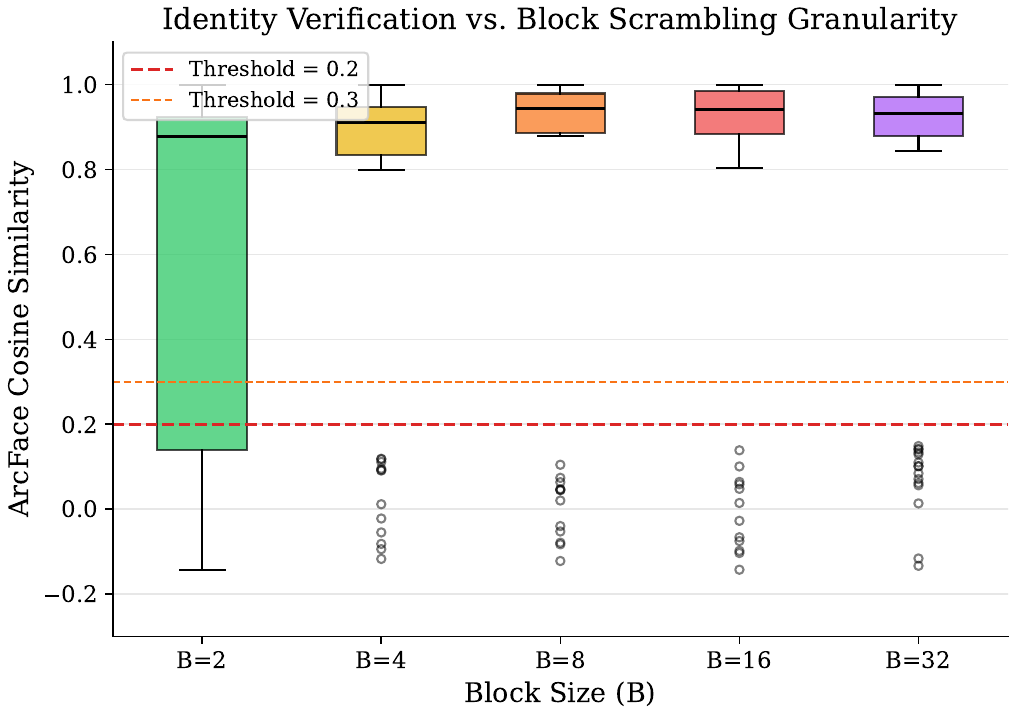}
    %\missing{FIGURE 6: Box plot. X-axis: Block size B (2, 4, 8, 16, 32). Y-axis: Cosine similarity. Dashed lines at 0.2 and 0.3 thresholds. The plot is generated by patch\_missing.py and saved as arcface\_block\_plot.png. Replace this placeholder with the actual figure file.}
    \caption{ArcFace \cite{deng2019arcface} cosine similarity between original and scrambled face crops as a function of block size $B$, computed over the subset of faces that remained detectable post-scrambling. The dashed lines indicate verification thresholds at 0.2 (conservative) and 0.3 (standard). Approximately 89\% of faces are not represented in this plot because they became undetectable after scrambling.}
    \label{fig:arcface_block_size}
\end{figure}

The lack of a clear trend across block sizes is attributable to the low resolution of UCF101 faces. At $224 \times 224$ frame resolution, face regions typically span 30--50 pixels per side; even a $B=32$ block covers the entire face in a single permutation unit. The 89\% face detection failure rate across all block sizes empirically validates the inclusion of $B \in \{4, 8, 16\}$ in the dataset: all three granularities provide strong identity privacy as measured by face detectability.

\subsection{Privacy Metric Summary}

% --- TABLE: Privacy metrics per tier ---
\begin{table}[t]
    \centering
    \caption{Privacy and utility metrics across tiers. SSIM and PSNR are computed within the ROI bounding box only. Face Detection Failure Rate is the percentage of originally-detectable faces that become undetectable after transformation. Face Detection Failure Rate is reported only for Tier 3 variants, where block scrambling is applied; Tier 1 and Tier 2 are marked with dashes because the ArcFace evaluation (Section \ref{subsec:block_size_experiment}) was conducted exclusively on scrambled outputs. $PU$ score is the composite Privacy-Utility metric (higher is better). All metrics are computed over 7,728 sampled frames.}
    \label{tab:privacy_metrics}
    \small
    \begin{tabular}{lccccc}
        \toprule
        \textbf{Tier} & \textbf{ROI-SSIM $\downarrow$} & \textbf{ROI-PSNR $\downarrow$} & \textbf{Face Fail \% $\uparrow$} & \textbf{Acc \% $\uparrow$} & \textbf{$PU$ $\uparrow$} \\
        \midrule
        Original        & 1.000 & $\infty$ & 0\%   & 88.8 & --- \\
        Tier1 (Blur)    & 0.764 & 34.88    & ---   & 66.5 & 0.177 \\
        Tier2 (Edge)    & 0.043 & 6.50     & ---   & 66.2 & 0.713 \\
        Tier3a ($B=16$) & 0.681 & 35.04    & 88.8\% & 66.2 & 0.238 \\
         Tier3b ($B=8$)  & 0.624 & 32.01    & 89.4\% & 64.4 & 0.273 \\
         Tier3c ($B=4$)  & 0.588 & 30.38    & 89.2\% & 63.1 & 0.293 \\
        \bottomrule
    \end{tabular}
\end{table}

Table~\ref{tab:privacy_metrics} presents the combined privacy-utility landscape. Tier~2 (Edge) achieves the highest $PU$ score (0.713) because it combines strong visual privacy (ROI-SSIM = 0.043) with competitive recognition accuracy (66.2\%). Among the Tier~3 variants, $B=4$ achieves the best $PU$ score (0.293) due to its lower SSIM despite a slight accuracy penalty. Tier~1 (Blur) has the lowest $PU$ score (0.177) because its relatively high SSIM (0.764) indicates weak visual privacy even though its recognition accuracy matches the other tiers.

The ROI-PSNR values for Tier~3 variants (30--35~dB) are counterintuitively high because PSNR measures pixel-level distortion: block scrambling rearranges blocks without altering pixel values within each block, resulting in many pixel pairs that happen to match their original positions. SSIM, which captures structural distortion, is the more informative metric for block-scrambling privacy.

% --- FIGURE: Privacy-Utility scatter plot ---
\begin{figure}[t]
    \centering
    \includegraphics[width=0.8\textwidth]{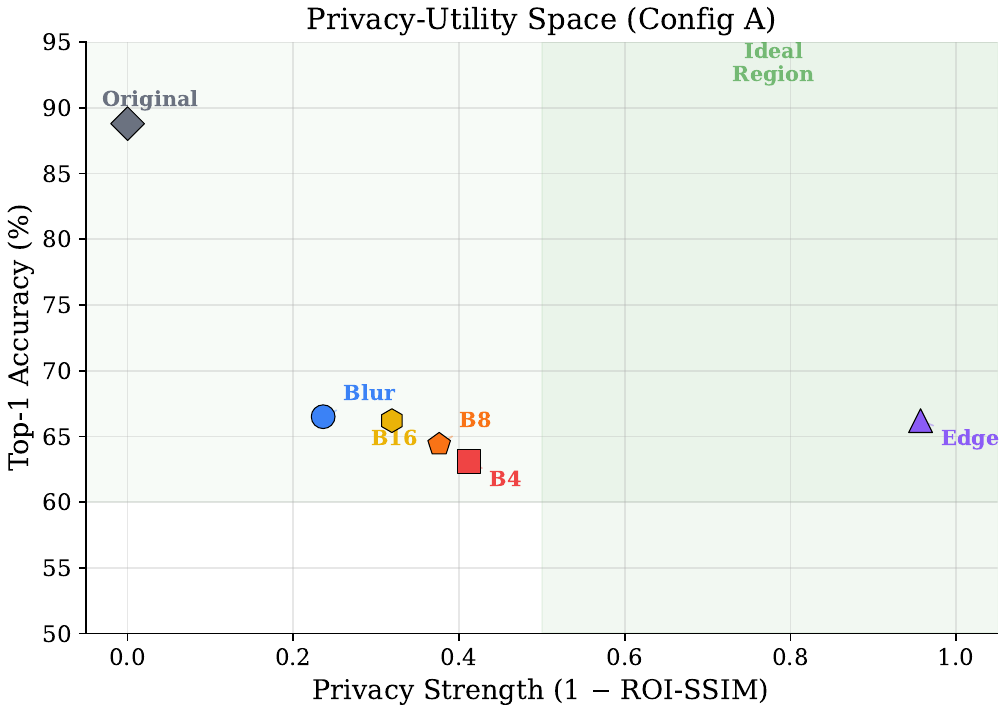}
    %\missing{FIGURE 7: Scatter plot. X-axis: Privacy (1 - ROI-SSIM). Y-axis: Utility (Top-1 Accuracy). Points: Original (0, 88.8), Blur (0.236, 66.5), Edge (0.957, 66.2), B4 (0.412, 63.1), B8 (0.376, 64.4), B16 (0.319, 66.2). Label each point. The ideal region is top-right.}
    \caption{Privacy-Utility space for the R3D-18 baseline on PrivHAR-Bench. Each point represents one tier under Config A. The x-axis measures privacy strength ($1 - \text{SSIM}$; higher is more private); the y-axis measures recognition utility. The top-right quadrant represents the ideal operating region: high privacy with high accuracy. Tier 2 (Edge) occupies the most favorable position.}
    \label{fig:pu_scatter}
\end{figure}

%!TEX root = Privacy_Dataset_Paper.tex

\section{Limitations and Future Work}
\label{sec:limitations}

\paragraph{Source resolution.} UCF101's native resolution of $320 \times 240$ is substantially lower than contemporary surveillance cameras (720p--1080p). The effectiveness of block scrambling is resolution-dependent: a $B=8$ block at $224 \times 224$ covers approximately $3.6\%$ of the frame width, whereas the same block at $1080 \times 1920$ covers less than $0.8\%$. Privacy metrics (SSIM, PSNR, ArcFace verification failure) computed on PrivHAR-Bench should not be directly extrapolated to higher-resolution deployments without re-evaluation. Future versions of this benchmark will target higher-resolution sources to address this gap.

\paragraph{Laboratory vs.\ in-the-wild conditions.} UCF101 videos are sourced from YouTube and exhibit variable lighting, camera angles, and recording quality. While this provides some naturalistic variation, it does not replicate the specific characteristics of fixed-position surveillance cameras: elevated mounting angles, wide-angle lens distortion, low-light conditions, and continuous multi-hour recording \cite{kansal2025implications}. A surveillance-specific benchmark requires purpose-collected data, which is outside the scope of this initial release.

\paragraph{Single-person assumption.} The current pipeline selects the single largest detected person per frame. Multi-person scenes (e.g., group activities, crowded environments) are not addressed. Classes involving person-person interaction are excluded. Extending the benchmark to multi-person privacy-preserving activity recognition is a distinct and open problem.

\paragraph{Temporal privacy.} The 32-frame temporal window reduces but does not eliminate the risk of gait-based re-identification. Research on gait recognition has demonstrated successful identification from short sequences under controlled conditions \cite{chao2019gaitset, fan2020gaitpart}. The current benchmark applies spatial privacy transformations only; temporal privacy (e.g., frame-order encryption, temporal downsampling) is not addressed. A formal temporal re-identification audit, training a model to classify actor identity rather than action class on encrypted clips, is planned for a future version and will quantify this residual risk.

\paragraph{Baseline training asymmetry.} The R3D-18 baseline on the Original tier was trained for 15 epochs, while all privacy tiers were trained for 5 epochs due to computational constraints (${\sim}$2.25 hours per epoch on an RTX~4070). This asymmetry slightly disadvantages the privacy tier results. The reported accuracy values should be interpreted as lower bounds; longer training would likely improve privacy-tier performance by 2--5 percentage points based on the convergence behavior observed during training.

\paragraph{Pose estimation quality.} The keypoints distributed in \texttt{/Estimated\_Poses} are produced by YOLOv8n-Pose \cite{jocher2023yolov8} and are \textit{estimated}, not ground truth. Joint confidence varies across classes and is lowest for actions involving self-occlusion (e.g., BabyCrawling) or fast motion (e.g., JumpRope). Users of the pose data should filter by confidence threshold appropriate to their application; we deliberately distribute raw confidence values rather than applying an arbitrary cutoff.

\paragraph{Class diversity.} The 15 selected classes are constrained by what UCF101 \cite{soomro2012ucf101} offers. Critical surveillance-relevant actions: falling, aggression, shoplifting, are absent because UCF101 does not contain reliable, unconstrained examples of these activities. PrivHAR-Bench evaluates privacy \textit{mechanisms} rather than surveillance \textit{applicability}. Extending the class list to include surveillance-specific activities from dedicated sources (e.g., NTU RGB+D 120 \cite{liu2019ntu} health-related classes) is planned for future releases.

\paragraph{Encryption method scope.} The current Tier~3 implements only block permutation scrambling \cite{qi2023privacy, ramasamy2019image}. Other cryptographic approaches: byte-level XOR encryption, selective frequency-domain encryption \cite{liu2021video}, format-preserving encryption, are not represented. Expanding the cryptographic tier to include multiple encryption methods would broaden the benchmark's audience but also increase storage and complexity substantially.

%!TEX root = Privacy_Dataset_Paper.tex

\section{Conclusion}
\label{sec:conclusion}

We presented PrivHAR-Bench, a multi-tier benchmark for evaluating the privacy-utility trade-off in video-based human activity recognition. By providing graduated privacy transformations, from spatial obfuscation to cryptographic block permutation, applied to a common set of 1,932 videos across 15 classes with standardized splits, lossless storage, and a fixed evaluation protocol, PrivHAR-Bench enables direct, reproducible comparison of privacy-preserving HAR methods for the first time.

Baseline experiments with R3D-18 confirm that the tier structure produces a measurable and interpretable degradation curve: recognition accuracy decreases from 88.8\% (clear) through 63--67\% (privacy tiers) to 53.5\% (encrypted with background removed). The context bias analysis demonstrates that background removal reduces accuracy by 10.9 percentage points under within-tier training and 29.6 percentage points under cross-domain evaluation, validating the necessity of the NoBG variant. The ArcFace analysis shows that block scrambling renders approximately 89\% of faces undetectable regardless of block size, providing empirical grounding for the privacy guarantees of the Tier~3 variants.

PrivHAR-Bench is a starting point, not a final product. Its limitations: source resolution, class diversity, single-person scope, spatial-only privacy, and training epoch asymmetry, are explicitly documented and define the roadmap for future extensions. We release the dataset, generation pipeline, and evaluation toolkit to enable the community to build on this foundation with full reproducibility.

The dataset is available at \url{10.5281/zenodo.19352048}. The generation code and evaluation toolkit are available at \url{https://github.com/SamarAnsariUK/PrivHAR-Bench}.

% Datasheet for Datasets (NeurIPS D&B requirement)
\newpage
%!TEX root = Privacy_Dataset_Paper.tex

\section*{Datasheet for PrivHAR-Bench}
\label{sec:datasheet}

This datasheet follows the framework proposed by Gebru et al.\ \cite{gebru2021datasheets} and is included as required by the NeurIPS Datasets and Benchmarks track.

\subsection*{Motivation}

\textbf{For what purpose was the dataset created?}
PrivHAR-Bench was created to provide a standardized benchmark for evaluating the trade-off between visual privacy and action recognition accuracy. No existing benchmark provides multiple graduated privacy tiers applied to the same source videos with a fixed evaluation protocol.

\textbf{Who created the dataset and on behalf of which entity?}
\texttt{Samar Ansari, University of Chester}

\textbf{Who funded the creation of the dataset?}
\texttt{Not Applicable}

\subsection*{Composition}

\textbf{What do the instances that comprise the dataset represent?}
Each instance is a 32-frame video clip of a single person performing one of 15 activity classes. Each clip exists in 9 versions (tiers), each representing a different level of visual privacy transformation.

\textbf{How many instances are there in total?}
1,932 source clips $\times$ 9 tiers = 17,388 total clip variants, comprising 556,416 individual frames.

\textbf{Does the dataset contain all possible instances or is it a sample?}
PrivHAR-Bench is a curated subset of UCF101 \cite{soomro2012ucf101}. It contains all clips from 15 selected classes.

\textbf{What data does each instance consist of?}
Each instance consists of: 32 lossless PNG frames at $224 \times 224$ resolution, per-frame bounding box coordinates and detection confidence, 17 estimated pose keypoints with per-joint confidence, and a class label.

\textbf{Is there a label or target associated with each instance?}
Yes. Each clip is labeled with one of 15 action classes. Labels are inherited from UCF101 annotations.

\textbf{Is any information missing from individual instances?}
Frames in which the person detector fails to detect a subject with sufficient confidence contain null bounding box and keypoint annotations. The detection rate per video is recorded in the metadata.

\textbf{Are relationships between individual instances made explicit?}
Yes. Each source clip generates one instance per tier. The \texttt{video\_id} field links all tier variants of the same source clip. Group membership is recorded for split integrity.

\textbf{Are there recommended data splits?}
Yes. Fixed group-based train/test splits are provided (1,452 train / 480 test). Users must not generate their own random splits, as this would introduce identity leakage and non-comparable results.

\textbf{Are there any errors, sources of noise, or redundancies?}
Estimated pose keypoints contain estimation noise, particularly for self-occluded joints. Per-joint confidence scores are provided to allow user-side filtering. The source UCF101 videos exhibit variable compression quality and occasional interlacing artifacts, which are inherited by PrivHAR-Bench.

\textbf{Is the dataset self-contained, or does it link to or otherwise rely on external resources?}
The dataset is self-contained. No external resources are required to use it. The generation pipeline, released separately, requires access to UCF101.

\textbf{Does the dataset contain data that might be considered confidential?}
No. UCF101 source videos are publicly available YouTube clips.

\textbf{Does the dataset contain data that, if viewed directly, might be offensive, insulting, threatening, or might otherwise cause anxiety?}
The Original (clear) tier contains unmodified human faces from UCF101. No content was flagged as offensive during curation. The privacy-transformed tiers obscure or destroy facial features.

\textbf{Does the dataset identify any subpopulations?}
No demographic annotations are provided. UCF101 does not include demographic metadata for its subjects.

\subsection*{Collection Process}

\textbf{How was the data associated with each instance acquired?}
The source videos were acquired from UCF101 \cite{soomro2012ucf101}, which collected them from YouTube. PrivHAR-Bench's privacy transformations were generated algorithmically using the pipeline described in Section~\ref{sec:dataset_generation}.

\textbf{What mechanisms or procedures were used to collect the data?}
Automated pipeline: YOLOv8n-Pose \cite{jocher2023yolov8} for ROI detection, OpenCV for blur and edge transformations, AES-CTR for block scrambling. All processing is deterministic and reproducible.

\textbf{If the dataset is a sample from a larger set, what was the sampling strategy?}
Classes were selected based on articulation diversity, minimal context/object dependency, and detection feasibility (see Section~\ref{subsec:class_selection}). All clips within selected classes were included.

\textbf{Who was involved in the data collection process?}
The dataset was generated programmatically. No manual annotation was performed beyond class selection validation.

\textbf{Over what timeframe was the data collected?}
The source UCF101 data was collected circa 2012. PrivHAR-Bench transformations were generated in March-April 2026.

\textbf{Were any ethical review processes conducted?}
%Ethical approval was obtained from the Science and Engineering Research Ethics Committee, Faculty of Science, Business and Enterprise, University of Chester . No human participants were recruited, no new data was collected, and all source material is derived from the publicly available UCF101 dataset. PrivHAR-Bench is derived entirely from UCF101, which consists of publicly available YouTube videos. No new human subjects data was collected, and no personally identifiable information beyond what appears in the source dataset is introduced.
Ethical approval was obtained from the Science and Engineering Research Ethics Committee, Faculty of Science, Business and Enterprise, University of Chester. PrivHAR-Bench is derived entirely from UCF101, which consists of publicly available YouTube videos. No human participants were recruited, no new data was collected, and no personally identifiable information beyond what appears in the source dataset is introduced.

\subsection*{Preprocessing / Cleaning / Labeling}

\textbf{Was any preprocessing/cleaning/labeling of the data done?}
Videos were clipped to 32 frames (center crop), resized to $224 \times 224$, and stored as lossless PNG. No manual cleaning was performed on the UCF101 labels, which are inherited as-is.

\textbf{Was the ``raw'' data saved in addition to the preprocessed/cleaned/labeled data?}
The Original tier contains the resized but otherwise unmodified frames. The raw UCF101 videos are available from the UCF101 distribution.

\textbf{Is the software that was used to preprocess/clean/label the data available?}
Yes. The complete generation pipeline is released at \url{https://github.com/SamarAnsariUK/PrivHAR-Bench}.

\subsection*{Uses}

\textbf{Has the dataset been used for any tasks already?}
Baseline experiments are reported in this paper (Section~\ref{sec:experiments}).

\textbf{Is there a repository that links to any or all papers or systems that use the dataset?}
Will be maintained at: \url{https://github.com/SamarAnsariUK/PrivHAR-Bench}

\textbf{What (other) tasks could the dataset be used for?}
Beyond privacy-preserving HAR, PrivHAR-Bench could be used for: evaluating robustness of video models to input degradation, studying the information content of edge/structural representations, benchmarking reconstruction attacks against visual encryption, and testing pose estimation under adversarial visual conditions.

\textbf{Is there anything about the composition of the dataset or the way it was collected and preprocessed/cleaned/labeled that might impact future uses?}
The source resolution ($320 \times 240$) limits the applicability of privacy metric findings to higher-resolution deployments. The single-person assumption limits use for multi-person activity recognition research.

\textbf{Are there tasks for which the dataset should not be used?}
PrivHAR-Bench should not be used as evidence that a privacy method is ``sufficient'' for real-world deployment. The benchmark evaluates relative performance across controlled conditions, not absolute privacy guarantees in operational environments.

\subsection*{Distribution}

\textbf{Will the dataset be distributed to third parties outside of the entity on behalf of which the dataset was created?}
Yes. The dataset is publicly available.

\textbf{How will the dataset be distributed?}
Via Zenodo as segmented ZIP archives, with a persistent DOI: \url{10.5281/zenodo.19352048}.

\textbf{When will the dataset be distributed?}
\texttt{1 April 2026}

\textbf{Will the dataset be distributed under a copyright or other intellectual property (IP) license?} CC-BY-NC-4.0
%\missing{Specify license. CC-BY-4.0 is recommended. Must be compatible with UCF101's terms of use.}

\textbf{Have any third parties imposed IP-based or other restrictions on the data associated with the instances?}
UCF101 is released for research purposes. PrivHAR-Bench inherits this restriction. Commercial use is not permitted.

\subsection*{Maintenance}

\textbf{Who will be supporting/hosting/maintaining the dataset?}
\texttt{Samar Ansari / University of Chester}

\textbf{How can the owner/curator/manager of the dataset be contacted?}
\texttt{m.ansari@chester.ac.uk}

\textbf{Is there an erratum?}
The \texttt{CHANGELOG.md} file in the dataset root documents all corrections. The \texttt{manifest.json} provides per-file SHA-256 hashes for integrity verification.

\textbf{Will the dataset be updated?}
Yes. Bug fixes and corrections will be released as patch versions (e.g., v1.0.1). Feature extensions (e.g., new source datasets, new tiers) will be released as minor or major versions. All published papers should cite the specific version used.

\textbf{Will older versions of the dataset continue to be supported/hosted/maintained?}
Yes. All released versions will remain available on Zenodo with their original DOI.

\textbf{If others want to extend/augment/build on/contribute to the dataset, is there a mechanism for them to do so?}
The generation pipeline is open-source. Users can apply it to other source datasets to produce compatible benchmarks. Contributions to the evaluation toolkit are welcome via the project repository.

\bibliographystyle{unsrt}
\bibliography{references}

\end{document}